\documentclass[11pt,a4paper]{article}
\usepackage[hyperref]{acl2021}
\usepackage{times}
\usepackage{makecell}
\usepackage{latexsym}
\usepackage{amsmath}
\usepackage{amsfonts}
\usepackage{hhline}
\usepackage{graphicx}
\usepackage{caption}
\usepackage{subcaption}

\usepackage{microtype}

\usepackage{xfrac}
\usepackage{colortbl}
\usepackage{booktabs}  % professional-quality tables
\usepackage{amsfonts}  % blackboard math symbols
\usepackage{nicefrac}
\usepackage{latexsym}
\usepackage{multirow}
\usepackage{amsmath}
\usepackage{tabularx}
\usepackage{makecell}
\usepackage{xcolor}

\definecolor{gred}{RGB}{219,68,55}
\definecolor{gblue}{RGB}{66,133,244}
\definecolor{gyellow}{RGB}{244,180,0}
\definecolor{ggreen}{RGB}{85,157,88}
\definecolor{ggrey}{RGB}{115,115,115}
\definecolor{na}{gray}{0.9}

\aclfinalcopy % Uncomment this line for the final submission

\title{ \textsc{G-Eval}: NLG Evaluation  using \textsc{Gpt}-4 with Better Human Alignment
}

\author{
  Yang Liu\quad Dan Iter\quad Yichong Xu 
  \\
  {\bf Shuohang Wang \quad Ruochen Xu\quad
  Chenguang Zhu} \\ \\
  Microsoft Cognitive Services Research
  \\ \emph{\{\textbf{yaliu10}, iterdan, yicxu, shuowa, ruox, chezhu\}@microsoft.com}}

\begin{document}

\maketitle

\begin{abstract}

The quality of texts generated by natural language generation (NLG) systems is hard to measure automatically. 
Conventional reference-based metrics, such as BLEU and ROUGE, have been shown to have relatively low correlation with human judgments, especially for tasks that require creativity and diversity. 
Recent studies suggest using large language models (LLMs) as reference-free metrics for NLG evaluation, which have the benefit of being applicable to new tasks that lack human references.
However, these LLM-based evaluators still have lower human correspondence than medium-size neural evaluators. 
In this work, we present \textsc{G-Eval}, a framework of using large language models with chain-of-thoughts (CoT) and a form-filling paradigm, to assess the quality of NLG outputs.  We experiment with two generation tasks, text summarization and dialogue generation. 
We show that \textsc{G-Eval} with GPT-4 as the backbone model achieves a Spearman  correlation of $0.514$ with human on summarization task, outperforming all previous methods by a large margin.
We also propose analysis on the behavior of LLM-based evaluators, and highlight the potential concern of LLM-based evaluators having a bias towards the LLM-generated texts. \footnote{\url{https://github.com/nlpyang/geval}}

\end{abstract}

\section{Introduction}

Evaluating the quality of natural language generation systems is a challenging problem even when large language models can generate high-quality and diverse texts that are often indistinguishable from human-written texts~\citep{gpt35}. 
Traditional automatic metrics, such as BLEU \cite{papineni2002bleu}, ROUGE \cite{lin2004rouge}, and METEOR \cite{banerjee2005meteor}, are widely used for NLG evaluation, but they have been shown to have relatively low correlation with human judgments, especially for open-ended generation tasks. Moreover, these metrics require associated reference output, which is costly to collect for new tasks.

Recent studies propose directly using LLMs as reference-free NLG evaluators \cite{fu2023GPTScore, wang2023chatgpt}. The idea is to use the LLMs to score the candidate output based on its generation probability without any reference target, under the assumption that the LLMs have learned to assign higher probabilities to high-quality and fluent texts. However, the validity and reliability of using LLMs as NLG evaluators have not been systematically investigated. In addition, meta-evaluations show that these LLM-based evaluators still have lower human correspondence than medium-size neural evaluators~\cite{zhong2022towards}.
Thus, there is a need for a more effective and reliable framework for using LLMs for NLG evaluation.

In this paper, we propose \textsc{G-Eval}, a framework of using LLMs  with chain-of-thoughts (CoT)~\cite{wei2022chain} to evaluate the quality of generated texts in a form-filling paradigm. 
By only feeding the Task Introduction and the Evaluation Criteria as a prompt, we ask LLMs to generate a CoT of detailed Evaluation Steps.
Then we  use the prompt along with the generated CoT to evaluate the NLG outputs.
The evaluator output is formatted as a form. Moreover, the probabilities of the output rating tokens can be used to refine the final metric.
We conduct extensive experiments on three meta-evaluation benchmarks of two NLG tasks: text summarization and dialogue generation. The results show that \textsc{G-Eval} can outperform existing NLG evaluators by a large margin in terms of correlation with human evaluations.
Finally, we conduct analysis on the behavior of LLM-based evaluators, and highlight the potential issue of LLM-based evaluator having a bias towards the LLM-generated texts.

\begin{figure*}
    \centering
    \includegraphics[width=1.0\linewidth]{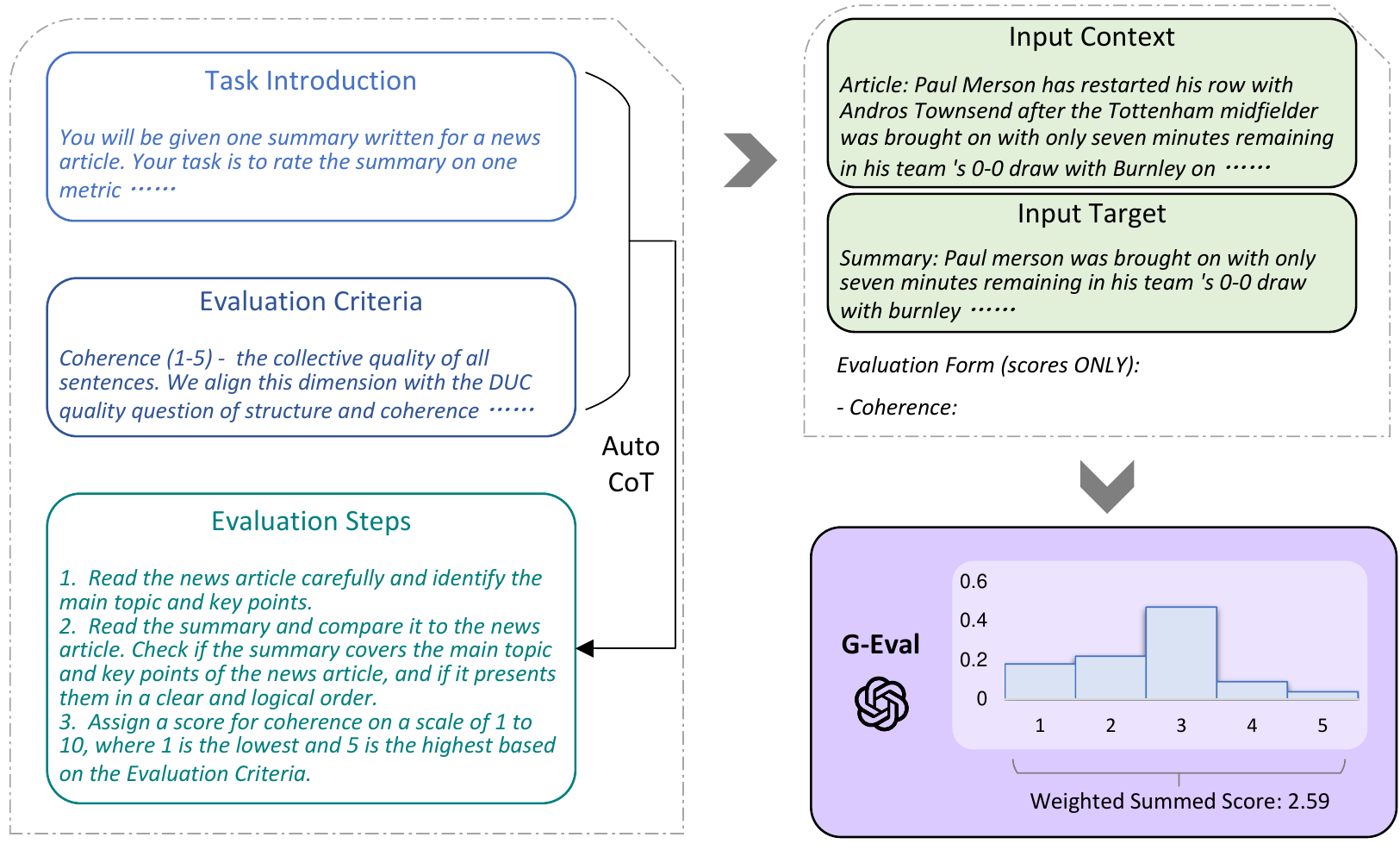}
    \caption{The overall framework of \textsc{G-Eval}. We first input Task Introduction and  Evaluation Criteria to the LLM, and ask it to generate a CoT of detailed Evaluation Steps. Then we  use the prompt along with the generated CoT to evaluate the NLG outputs in a form-filling paradigm. Finally, we use the probability-weighted  summation of the output scores as the final score.
    }
    \label{fig:model}
\end{figure*}

To summarize, our main contributions in this paper are:

\begin{enumerate}
    
\item LLM-based metrics generally outperform reference-based and reference-free baseline metrics in terms of correlation with human quality judgments, especially for open-ended and creative NLG tasks, such as dialogue response generation.
    
\item LLM-based metrics are sensitive to the instructions and prompts, and chain-of-thought can improve the performance of LLM-based evaluators by providing more context and guidance.

\item LLM-based metrics can provide a more fine-grained continuous score by re-weighting the discrete scores by their respective token probabilities.
    
\item LLM-based metrics have a potential issue of preferring  LLM-generated texts over human-written texts, which may lead to the self-reinforcement of LLMs if LLM-based metrics are used as the reward signal for improving themselves.

\end{enumerate}

\section{Method}

\textsc{G-Eval} is a prompt-based evaluator with three main components: 1) a prompt that contains the definition of the evaluation task and the desired evaluation criteria, 2) a chain-of-thoughts (CoT) that is a set of intermediate instructions generated by the LLM describing the detailed evaluation steps, and 3) a scoring function that calls LLM and calculates the score based on the probabilities of the return tokens.

\paragraph{Prompt for NLG Evaluation} The prompt is a natural language instruction that defines the evaluation task and the desired evaluation criteria. For example, for text summarization, the prompt can be:
\begin{quote}
\emph{You will be given one summary written for a news article.
Your task is to rate the summary on one metric.}

\emph{Please make sure you read and understand these instructions carefully. Please keep this document open while reviewing, and refer to it as needed.}
\end{quote}

The prompt should also contain customized  evaluation criteria for different NLG tasks and, such as coherence, conciseness, or grammar. For example, for evaluating coherence in text summarization, we add the following content to the prompt:

\begin{quote}
\emph{Evaluation Criteria:}

\emph{Coherence (1-5) -  the collective quality of all sentences. We align this dimension with the DUC quality question of structure and coherence whereby "the summary should be well-structured and well-organized. The summary should not just be a heap of related information, but should build from sentence to sentence to a coherent body of information about a topic."
}
\end{quote}

\paragraph{Auto Chain-of-Thoughts for NLG Evaluation}  The chain-of-thoughts (CoT) is a sequence of intermediate representations that are generated by the LLM during the text generation process. For evaluation tasks, some criteria need a more detailed evaluation instruction beyond the simple definition, and it is time-consuming to manually design such evaluation steps for each task. We find that LLM can generate such evaluation steps by itself. The CoT can provide more context and guidance for the LLM to evaluate the generated text, and can also help to explain the evaluation process and results.  For example, for evaluating coherence in text summarization, we add a line of \emph{``Evaluation Steps:"} to the prompt and let  LLM to generate the following CoT automatically:

\begin{quote}
\emph{1.  Read the news article carefully and identify the main topic and key points.}

\emph{2.  Read the summary and compare it to the news article. Check if the summary covers the main topic and key points of the news article, and if it presents them in a clear and logical order.}

\emph{3.  Assign a score for coherence on a scale of 1 to 5, where 1 is the lowest and 5 is the highest based on the Evaluation Criteria.}
\end{quote}

\paragraph{Scoring Function} The scoring function calls the LLM with the designed prompt, auto CoT, the input context and the target text that needs to be evaluated. Unlike GPTScore \cite{fu2023GPTScore} which uses the conditional probability of generating the target text as an evaluation metric, \textsc{G-Eval} directly performs the evaluation task with a form-filling paradigm.  For example, for evaluating coherence in text summarization, we concatenate the prompt, the CoT, the news article, and the summary, and then call the LLM to output a score from 1 to 5 for each evaluation aspect, based on the defined criteria.

However, we notice this direct scoring function has two issues:

\begin{enumerate}
    \item For some evaluation tasks, one digit usually dominates the distribution  of the scores, such as 3 for a 1 - 5 scale.  This may lead to the low variance of the scores and the low correlation with human judgments. 
    \item LLMs usually only output integer scores, even when the prompt explicitly requests decimal values. This leads to many ties in evaluation scores which do not capture  the subtle difference between generated texts. 
\end{enumerate}

To address these issues, we propose using the probabilities of output tokens from LLMs to normalize the scores and take their weighted summation as the final results. Formally, given a set of scores (like from 1 to 5) predefined in the prompt 
$S=\{s_1, s_2, ..., s_n\}$, the probability of each score $p(s_i)$ is calculated by the LLM, and the final score is:

\begin{equation}
    score = \sum_{i=1}^n p(s_i) \times s_i
\end{equation}

This method obtains more fine-grained, continuous scores that better reflect the quality and diversity of the generated texts.

\begin{table*}[!htbp]
\begin{tabular}{l|cc|cc|cc|cc|cc}
\multicolumn{1}{c|}{\multirow{2}[1]{*}{\textbf{Metrics}}} & \multicolumn{2}{c|}{\textbf{Coherence}}
 & \multicolumn{2}{c|}{\textbf{Consistency}} & \multicolumn{2}{c|}{\textbf{Fluency}} & \multicolumn{2}{c|}{\textbf{Relevance}} & \multicolumn{2}{c}{\textbf{AVG}}  \\ 
            & $\rho$         & $\tau$         & $\rho$          & $\tau$         & $\rho$         & $\tau$         & $\rho$         & $\tau$         & $\rho$         & $\tau$         \\ \hline

ROUGE-1     & 0.167          & 0.126          & 0.160           & 0.130          & 0.115          & 0.094          & 0.326          & 0.252          & 0.192          & 0.150          \\
ROUGE-2     & 0.184          & 0.139          & 0.187           & 0.155          & 0.159          & 0.128          & 0.290          & 0.219          & 0.205          & 0.161          \\
ROUGE-L     & 0.128          & 0.099          & 0.115           & 0.092          & 0.105          & 0.084          & 0.311          & 0.237          & 0.165          & 0.128          \\ \hline
BERTScore   & 0.284          & 0.211          & 0.110           & 0.090          & 0.193          & 0.158          & 0.312          & 0.243          & 0.225          & 0.175          \\
MOVERSscore  & 0.159          & 0.118          & 0.157           & 0.127          & 0.129          & 0.105          & 0.318          & 0.244          & 0.191          & 0.148          \\
BARTScore & 0.448 & 0.342 & 0.382 & 0.315 & 0.356 & 0.292 & 0.356 & 0.273 & 0.385 & 0.305 \\
UniEval     & 0.575          & 0.442          & 0.446           & 0.371          & 0.449          & 0.371          & 0.426          & 0.325          & 0.474          & 0.377          \\ \hline
GPTScore&0.434&--&0.449&--&0.403&--&0.381&--&0.417&--
\\
\textsc{G-Eval}-3.5 & 0.440          & 0.335          & 0.386           & 0.318          & 0.424          & 0.347          & 0.385          & 0.293          & 0.401          & 0.320          \\

\ \ \ - Probs & 0.359 & \textit{0.313} & 0.361 & \textit{0.344} & 0.339 & \textit{0.323} & 0.327 & \textit{0.288} & 0.346 & \textit{0.317}\\

\textsc{G-Eval}-4   & \textbf{0.582} & \textbf{0.457} & \textbf{0.507}  & \textbf{0.425} & \textbf{0.455} & \textbf{0.378} & \textbf{0.547} & \textbf{0.433} & \textbf{0.514} & \textbf{0.418}\\
\ \ \ - Probs &	0.560&	\textit{0.472}&	0.501&	\textit{0.459}&	0.438&	\textit{0.408}&	0.511&	\textit{0.444}&	0.502&	\textit{0.446}\\
\ \ \ - CoT &	0.564&	0.454&	0.493&	0.413&	0.403&	0.334&	0.538&	0.427&	0.500&	0.407

\end{tabular}
\caption{Summary-level Spearman ($\rho$) and Kendall-Tau ($\tau$) correlations of different metrics on SummEval benchmark. \textsc{G-Eval} without probabilities (\textit{italicized}) should not be considered  as a fair comparison to other metrics on $\tau$, as it leads to many ties in the scores. This results in a higher Kendall-Tau correlation, but it does not fairly reflect the true evaluation ability. More details are in Section~\ref{sec:analysis}.
}
\label{tab:summ}
\end{table*}

\section{Experiments}
Following~\citet{zhong2022towards}, we meta-evaluate our evaluator on three benchmarks, SummEval, Topical-Chat and QAGS, of two NLG tasks, summarization and dialogue response generation.

\subsection{Implementation Details}
We use OpenAI's GPT family as our LLMs, including GPT-3.5 (text-davinci-003) and GPT-4. For GPT-3.5, we set decoding temperature to 0 to increase the model's determinism. For GPT-4, as it does not support the output of token probabilities, we set `$n=20, temperature=1, top\_p=1$' to sample 20 times to estimate the token probabilities. 
We use \textsc{G-Eval}-4 to indicate \textsc{G-Eval}  with GPT-4 as the backbone model, and  \textsc{G-Eval}-3.5 to indicate \textsc{G-Eval}  with GPT-3.5 as the backbone model. Example prompts for each task are provided in the Appendix.

\subsection{Benchmarks}

We adopt three meta-evaluation benchmarks  to measure the correlation between \textsc{G-Eval} and human judgments.
\\\\
\textbf{SummEval}~\cite{fabbri2021summeval} is a benchmark that compares different evaluation methods for summarization. It gives human ratings for four aspects of each summary: \texttt{fluency}, \texttt{coherence}, \texttt{consistency} and \texttt{relevance}. It is built on the CNN/DailyMail dataset~\cite{hermann2015teaching}
\\\\
\textbf{Topical-Chat}~\cite{mehri2020usr} is a testbed for meta-evaluating different evaluators on dialogue response generation systems that use knowledge. We follow \cite{zhong2022towards} to use its human ratings on four aspects: \texttt{naturalness}, \texttt{coherence}, \texttt{engagingness} and \texttt{groundedness}.
\\\\
\textbf{QAGS}~\cite{wang2020asking} is a benchmark for evaluating hallucinations in the summarization task. It aims to measure the \texttt{consistency} dimension of summaries on two different summarization datasets.

\subsection{Baselines}
We evaluate \textsc{G-Eval} against various evaluators that achieved state-of-the-art performance. 

\textbf{BERTScore}~\cite{zhang2019bertscore} measures the similarity between two texts based on the contextualized embedding from BERT~\cite{devlin2019bert}.

\textbf{MoverScore}~\cite{zhao2019MoverScore} improves BERTScore by adding soft alignments and new aggregation methods to obtain a more robust similarity measure.

\textbf{BARTScore}~\cite{yuan2021bartscore} is a unified evaluator which evaluate with the average likelihood of the pretrained encoder-decoder model, BART \cite{DBLP:conf/acl/LewisLGGMLSZ20}. It can predict different scores depending on the formats of source and target.

\textbf{FactCC} and \textbf{QAGS}~\cite{kryscinski2020evaluating, wang2020asking}  are two evaluators that measure the factual consistency of generated summaries. FactCC is a BERT-based classifier that predicts whether a summary is consistent with the source document. QAGS is a question-answering based evaluator that generates questions from the summary and checks if the answers can be found in the source document.

\textbf{USR}~\cite{mehri2020usr} is evaluator that  assess dialogue response generation from different perspectives. It has several versions that assign different scores to each target response.

\textbf{UniEval}~\cite{zhong2022towards} is a unified evaluator that can evaluate different aspects of text generation as QA tasks. It uses a pretrained T5 model \cite{raffel2020exploring} to encode the evaluation task, source and target texts as questions and answers, and then computes the QA score as the evaluation score. It can also handle different evaluation tasks by  changing the question format.

\textbf{GPTScore}~\cite{fu2023GPTScore} is a new framework that evaluates texts with generative pre-training models like GPT-3. It assumes that  a generative pre-training
model will assign a higher probability of high-quality generated text following a given instruction and context. Unlike \textsc{G-Eval}, GPTScore formulates the evaluation task as a conditional generation problem instead of a form-filling problem.

\begin{table*}[]
\begin{tabular}{l|cc|cc|cc|cc|cc}

\multicolumn{1}{c|}{\multirow{2}[1]{*}{\textbf{Metrics}}} & \multicolumn{2}{c|}{\textbf{Naturalness}}
 & \multicolumn{2}{c|}{\textbf{Coherence}} & \multicolumn{2}{c|}{\textbf{Engagingness}} & \multicolumn{2}{c|}{\textbf{Groundedness}} & \multicolumn{2}{c}{\textbf{AVG}}  \\ 
            & $r$ & $\rho$ & $r$ & $\rho$ & $r$ & $\rho$ & $r$ & $\rho$ & $r$ & $\rho$      \\ \hline

ROUGE-L&0.176&0.146&0.193&0.203&0.295&0.300&0.310&0.327&0.243&0.244\\
BLEU-4&0.180&0.175&0.131&0.235&0.232&0.316&0.213&0.310&0.189&0.259\\
METEOR&0.212&0.191&0.250&0.302&0.367&0.439&0.333&0.391&0.290&0.331\\
BERTScore&0.226&0.209&0.214&0.233&0.317&0.335&0.291&0.317&0.262&0.273\\\hline
USR & 0.337 & 0.325 & 0.416 & 0.377 & 0.456 & 0.465 & 0.222 & 0.447 & 0.358 & 0.403\\
UniEval          & 0.455           & 0.330          & 0.602          & 0.455          & 0.573           & 0.430           & 0.577           & 0.453           & 0.552          & 0.417          \\ \hline
\textsc{G-Eval}-3.5 &0.532 & 0.539 & 0.519 & 0.544 & \textbf{0.660} & \textbf{0.691} & \textbf{0.586} & 0.567 & 0.574 & 0.585     \\

\textsc{G-Eval}-4  &\textbf{0.549} & \textbf{0.565} & \textbf{0.594} & \textbf{0.605} & 0.627 &0.631 &0.531 & 0.551 & \textbf{0.575} & \textbf{0.588}\\

\end{tabular}
\caption{Turn-level Spearman ($\rho$) and Kendall-Tau ($\tau$) correlations of different metrics on Topical-Chat benchmark. }
\label{tab:dialogue}
\end{table*}

\subsection{Results for Summarization}
We adopt the same approach as \citet{zhong2022towards} to evaluate different summarization metrics using summary-level Spearman and Kendall-Tau correlation. The first part of Table~\ref{tab:summ} shows the results of metrics that compare the semantic similarity between the model output and the reference text. These metrics perform poorly on most dimensions. The second part shows the results of metrics that use neural networks to learn from human ratings of summary quality. These metrics have much higher correlations than the similarity-based metrics, suggesting that they are more reliable for summarization evaluation.

In the last part of Table~\ref{tab:summ} which corresponds to GPT-based evaluators, GPTScore also uses GPTs for evaluating summarization texts, but relies on GPT's conditional probabilities of the given target.
\textsc{G-Eval} substantially surpasses all previous state-of-the-art evaluators on the SummEval benchmark.
\textsc{G-Eval}-4 achieved much higher human correspondence compared with \textsc{G-Eval}-3.5 on both Spearman and Kendall-Tau  correlation, which indicates that the larger model size of GPT-4 is beneficial for summarization evaluation. 
\textsc{G-Eval} also outperforms GPTScore on several dimension, demonstrating the effectiveness of the simple form-filling paradigm. 

\subsection{Results for Dialogue Generation}

We use the Topical-chat benchmark from \citet{mehri2020usr} to measure how well different evaluators agree with human ratings on the quality of dialogue responses. We calculate the Pearson and Spearman correlation for each turn of the dialogue. Table~\ref{tab:dialogue} shows that  similarity-based metrics  have good agreement with humans on how \texttt{engaging} and \texttt{grounded} the responses are, but not on the other aspects. 
With respect to the learning-based evaluators, before \textsc{G-Eval}, UniEval predicts scores that are most consistent with human judgments across all aspects.

As shown in the last part, 
\textsc{G-Eval}  also substantially surpasses all previous state-of-the-art evaluator on the Topical-Chat benchmark.
Notably, the \textsc{G-Eval}-3.5 can achieve similar results with \textsc{G-Eval}-4. This indicates that this benchmark is  relatively easy for the \textsc{G-Eval} model. 

\subsection{Results on Hallucinations}
\label{sec:qags}

Advanced NLG models  often produce text that does not match the context input~\cite{cao2018faithful}, and recent studies find even powerful LLMs also suffer from the problem of hallucination. This motivates recent research to design evaluators for measuring the \texttt{consistency} aspect in summarization~\cite{kryscinski2020evaluating, wang2020asking, cao2020factual, durmus2020feqa}.
We test the QAGS meta-evaluation benchmark, which includes two different summarization datasets: CNN/DailyMail and XSum~\cite{narayan2018don}
Table~\ref{tab:qags} shows that BARTScore performs well on the more extractive subset (QAGS-CNN), but has low correlation on the more abstractive subset (QAGS-Xsum).
UniEval has good correlation on both subsets of the data.

On average, \textsc{G-Eval}-4 outperforms all state-of-the-art evaluators  on QAGS, with a large margin on QAGS-Xsum.
\textsc{G-Eval}-3.5, on the other  hand, failed to perform well on this benchmark, which indicates that the consistency aspect is sensitive to the LLM's capacity. This result is consistent with Table~\ref{tab:summ}.

\section{Analysis}
\label{sec:analysis}
\paragraph{Will \textsc{G-Eval} prefer LLM-based outputs?}

One concern about using LLM as an evaluator is  that it may prefer the outputs generated by the LLM itself, rather than the high-quality human-written texts. To investigate this issue, we conduct an experiment on the summarization task, where we compare the evaluation scores of the LLM-generated and the human-written summaries. We use the dataset collected in \citet{https://doi.org/10.48550/arxiv.2301.13848}, where they first ask freelance writers to write high-quality summaries for news articles, and then ask annotators to compare human-written summaries  and LLM-generated summaries (using GPT-3.5, text-davinci-003).

The dataset can be divided in three  categories: 1) human-written summaries that are rated \emph{higher} than GPT-3.5 summaries by human judges, 2) human-written summaries that are rated  \emph{lower}  than GPT-3.5 summaries by human judges, and 3) human-written summaries and GPT-3.5 summaries are rated  \emph{equally} good by human judges. We use \textsc{G-Eval}-4 to evaluate the summaries in each category, and compare the averaged scores. \footnote{We use \textsc{G-Eval}-4 in this experiment, because its superiority in evaluating summarization tasks. Although it has different distribution with with GPT-3.5, the two LLMs should  share similar behaviors  in terms of text generation.}

\begin{figure}
    \centering
    \includegraphics[width=1.0\linewidth]{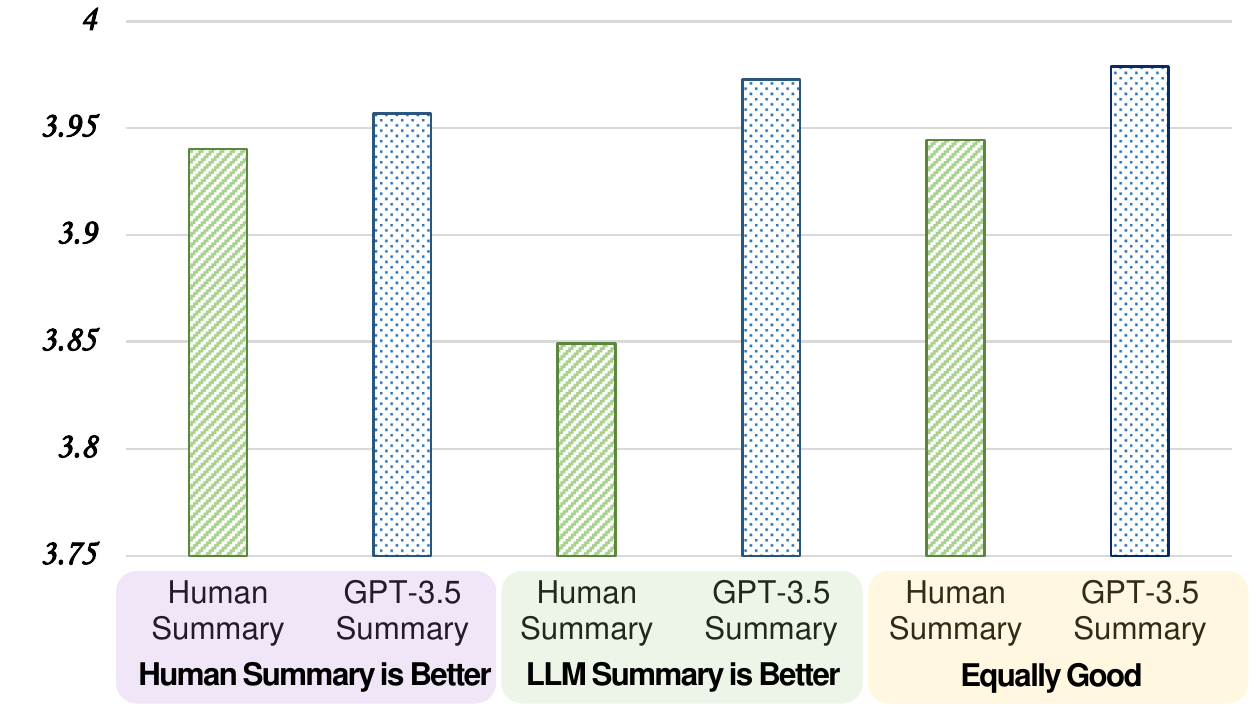}
    \caption{
    Averaged \textsc{G-Eval}-4's scores for human-written summaries and GPT-3.5 summaries, divided by human judges' preference. }
    \label{fig:ty}
\end{figure}

  The results are shown in Figure~\ref{fig:ty}.   We can see that, \textsc{G-Eval}-4 assigns higher scores to human-written summaries when human judges also prefer human-written summaries, and assigns lower scores when human judges prefer GPT-3.5 summaries. 
  However, \textsc{G-Eval}-4 always gives higher scores to GPT-3.5 summaries than human-written summaries, even when human judges prefer human-written summaries. 
We propose two potential reasons for this phenomenon: 

\begin{enumerate}
    \item  NLG outputs from high-quality systems are in natural difficult to evaluate. The authors of the original paper found that inter-annotator agreement on judging human-written  and LLM-generated summaries is very low, with Krippendorff's alpha at 0.07.
    \item  \textsc{G-Eval} may have a  bias towards the LLM-generated summaries  because the model could share the same concept of evaluation criteria during generation  and evaluation. 
\end{enumerate}

  Our work should be considered as  a preliminary study on this issue, and more research is needed to fully understand the behavior of LLM-based evaluators to reduce its inherent bias towards LLM-generated text.
  We highlight this concern in the context that LLM-based evaluators may lead to self-reinforcement of LLMs if the evaluation score is used as a reward signal for further tuning. And this could result in the over-fitting of the LLMs to their own evaluation criteria, rather than the true evaluation criteria of the NLG tasks.

\begin{table*}[t]
\center 
\begin{tabular}{l|ccc|ccc|ccc}
\multicolumn{1}{c|}{\multirow{2}[1]{*}{\textbf{Metrics}}} & \multicolumn{3}{c|}{\textbf{QAGS-CNN}}
 & \multicolumn{3}{c|}{\textbf{QAGS-XSUM}} & \multicolumn{3}{c}{\textbf{Average}}  \\ 
 
 & $r$ & $\rho$ & $\tau$ &  $r$ & $\rho$ & $\tau$ & $r$ & $\rho$ & $\tau$ \\ \hline

ROUGE-2 & 0.459 & 0.418 & 0.333 & 0.097 & 0.083 & 0.068 &  0.278 &  0.250 &  0.200  \\

ROUGE-L & 0.357 & 0.324 & 0.254 & 0.024 & -0.011 & -0.009 &  0.190 &  0.156 &  0.122  \\ \hline

BERTScore & 0.576 & 0.505 & 0.399 & 0.024 & 0.008 & 0.006 &  0.300 &  0.256 &  0.202  \\

MoverScore & 0.414 & 0.347 & 0.271 & 0.054 & 0.044 & 0.036 &  0.234 &  0.195 &  0.153  \\

FactCC & 0.416 & 0.484 & 0.376 & 0.297 & 0.259 & 0.212 &  0.356 &  0.371 &  0.294  \\

QAGS & 0.545 & - & - & 0.175 & - & - &  0.375 &  - &  - \\

BARTScore & \textbf{0.735} & 0.680 & 0.557 & 0.184 & 0.159 & 0.130 &  0.459 &  0.420 &  0.343  \\

CTC  & 0.619 & 0.564 & 0.450 & 0.309 & 0.295 & 0.242 &  0.464 &  0.430 &  0.346  \\

UniEval & 0.682 & 0.662 & 0.532 & 0.461 & 0.488 & 0.399 &  0.571 &  0.575 &  0.465  \\\hline
\textsc{G-Eval-3.5} & 0.477 &  0.516 & 0.410 &0.211 & 0.406  & 0.343   &  0.344	& 0.461	& 0.377  \\
\textsc{G-Eval-4} &0.631 & \textbf{0.685}  & \textbf{0.591} & \textbf{0.558} &  \textbf{0.537}  & \textbf{0.472}   & \textbf{0.599}	&  \textbf{0.611}	&  \textbf{0.525} \\

\end{tabular}
\caption{Pearson ($r$), Spearman ($\rho$) and Kendall-Tau ($\tau$) correlations of different metrics on QAGS benchmark.}
\label{tab:qags}
\end{table*}

\paragraph{The Effect of Chain-of-Thoughts} We compare the performance of \textsc{G-Eval} with and without chain-of-thoughts (CoT) on the SummEval benchmark. Table~\ref{tab:summ} shows that \textsc{G-Eval}-4 with CoT has higher correlation than \textsc{G-Eval}-4 without CoT on all dimensions, especially for \texttt{fluency}. This suggests that CoT can provide more context and guidance for the LLM to evaluate the generated text, and can also help to explain the evaluation process and results. 

\paragraph{The Effect of Probability Normalization} We compare the performance of \textsc{G-Eval} with and without probability normalization on the SummEval benchmark. Table~\ref{tab:summ} shows that,  on Kendall-Tau correlation, \textsc{G-Eval}-4 with probabilities is inferior  to \textsc{G-Eval}-4 without probabilities on SummEval. We believe this is related to the calculation of Kendall-Tau correlation, which is based on the number of concordant and discordant pairs. Direct scoring without probabilities can lead to many ties,  which are not counted as either concordant or discordant. This may result in a higher Kendall-Tau correlation, but it does not reflect the model's true capacity of evaluating the generated texts. On the other hand, probability normalization can obtain more fine-grained, continuous scores that better capture the subtle difference between generated texts. This is reflected by the higher Spearman correlation of \textsc{G-Eval}-4 with probabilities, which is based on the rank order of the scores.

\paragraph{The Effect of Model Size} We compare the performance of \textsc{G-Eval} with different model sizes on the SummEval and QAGS benchmarks. Table~\ref{tab:summ} and Table~\ref{tab:qags} show that \textsc{G-Eval}-4 has higher correlation than \textsc{G-Eval}-3.5 on most dimensions and datasets, except for \texttt{engagingness} and \texttt{groundedness} on the Topical-Chat benchmark. This demonstrates that larger model size can improve the performance of \textsc{G-Eval}, especially for more challenging and complex evaluation tasks, such as \texttt{consistency} and \texttt{relevance}.

\section{Related Work}

\paragraph{Ngram-based Metrics}
Ngram-based metrics refer to the scores for evaluating the NLG models by measuring the lexical overlap between a generated text and a reference text.
BLEU~\cite{papineni2002bleu} is the most widely used metric for machine translation evaluation, which calculates the geometric mean of modified n-gram precision and a brevity penalty. 
ROUGE~\cite{lin2004rouge} is a recall-oriented metric for summarization evaluation, which measures the n-gram overlap between a generated summary and a set of reference summaries. 
It has been  shown that more than 60\% of recent papers on NLG only rely on ROUGE or BLEU to evaluate their systems~\cite{kasai2021bidimensional}. However, these metrics fail to measure content quality~\cite{reiter2009investigation} or capture  syntactic errors~\cite{stent2005evaluating}, and therefore do not reflect the reliability of NLG systems accurately.

\paragraph{Embedding-based Metrics}
Embedding-based metrics refer to the scores for evaluating the NLG models by measuring the semantic similarity between a generated text and a reference text based on the word or sentence embeddings. 
WMD~\cite{kusner2015word} is a metric that measures the distance between two texts based on the word embeddings. 
BERTScore~\cite{zhang2019bertscore} measures the similarity between two texts based on the contextualized embedding from BERT~\cite{devlin2019bert}. 
MoverScore~\cite{zhao2019MoverScore} improves BERTScore by adding soft alignments and new aggregation methods to obtain a more robust similarity measure. 
\cite{clark2019sentence}  propose a metric that  evaluates multi-sentence texts 
by computing the similarity between the generated text and the reference text based on the sentence embeddings.

\paragraph{Task-specific Evaluators}
Task-specific metrics refer to the scores for evaluating the NLG models by measuring the quality of the generated texts based on the specific task requirements.  For example, summarization tasks need to assess the \texttt{consistency} of the generated summaries~\cite{kryscinski2020evaluating, wang2020asking, cao2020factual, durmus2020feqa}, and dialogue response generation tasks need to assess the \texttt{coherence} of the generated responses~\cite{dziri2019evaluating, ye2021towards}. However, these metrics are not generalizable to other NLG tasks, and they are not able to measure the overall quality of the generated texts.

\paragraph{Unified Evaluators}
Recently,  some evaluators have been developed to assess text quality from multiple dimensions by  varying the input and output contents~\cite{yuan2021bartscore} or the model variants~\cite{mehri2020usr} they use. UniEval  \cite{zhong2022towards} is a unified evaluator that can evaluate different aspects of text generation as QA tasks. By  changing the question format, it can handle different evaluation tasks.

\paragraph{LLM-based Evaluators}
\citet{fu2023GPTScore} propose GPTScore, a new framework that evaluated texts with generative pre-training models like GPT-3. It assumes that  a generative pre-training
model will assign a higher probability of high-quality generated text following a given instruction and context. 
\citet{wang2023chatgpt} conduct a preliminary survey of using ChatGPT as a NLG evaluator.
\citet{kocmi2023large} proposed to use GPT models for evaluating machine translation tasks.

\section{Conclusion}
In this paper, we propose \textsc{G-Eval}, a framework of using LLM with chain-of-thoughts (CoT) to evaluate the quality  of generated texts. We conduct extensive experiments on two NLG tasks, text summarization and dialogue generation, and show that \textsc{G-Eval} can outperform state-of-the-art evaluators and achieve higher human correspondence. We also propose preliminary  analysis on the behavior of LLM-based evaluators, and highlight the potential issue of LLM-based evaluator having a bias 
towards the LLM-generated texts. We hope our work can inspire more research on using LLMs for NLG evaluation, and also raise awareness of the potential risks and challenges of using LLMs as evaluators.

\newpage 

\bibliography{acl2021}
\bibliographystyle{acl_natbib}

\newpage

\onecolumn
\appendix

\section{Example Prompts}

\paragraph{Evaluate Coherence in the Summarization Task}

\begin{quote}
{\itshape
You will be given one summary written for a news article.

Your task is to rate the summary on one metric.

Please make sure you read and understand these instructions carefully. Please keep this document open while reviewing, and refer to it as needed.
\\
\\
Evaluation Criteria:

Coherence (1-5) -  the collective quality of all sentences. We align this dimension with the DUC quality question of structure and coherence whereby "the summary should be well-structured and well-organized. The summary should not just be a heap of related information, but should build from sentence to sentence to a coherent body of information about a topic."
\\
\\
Evaluation Steps:

1.  Read the news article carefully and identify the main topic and key points.

2.  Read the summary and compare it to the news article. Check if the summary covers the main topic and key points of the news article, and if it presents them in a clear and logical order.

3.  Assign a score for coherence on a scale of 1 to 5, where 1 is the lowest and 5 is the highest based on the Evaluation Criteria.
\\
\\
Example:

Source Text: 

\{\{Document\}\}

Summary: 

\{\{Summary\}\}
\\
\\
Evaluation Form (scores ONLY):

- Coherence:
}
\end{quote}

\paragraph{Evaluate Engagingness in the Dialogue Generation Task}

\begin{quote}
{\itshape
You will be given a conversation between two individuals. You will then be given one potential response for the next turn in the conversation. The response concerns an interesting fact, which will be provided as well. 

Your task is to rate the responses on one metric.

Please make sure you read and understand these instructions carefully. Please keep this document open while reviewing, and refer to it as needed.
\\
\\
Evaluation Crieteria:

Engagingness (1-3) Is the response dull/interesting?

- A score of 1 (dull) means that the response is generic and dull.

- A score of 2 (somewhat interesting) means the response is somewhat interesting and could engage you in the conversation (e.g., an opinion, thought)

- A score of 3 (interesting) means the response is very interesting or presents an interesting fact
\\
\\
Evaluation Steps:

1.  Read the conversation, the corresponding fact  and the response carefully.

2.   Rate the response on a scale of 1-3 for engagingness, according to the criteria above.

3.   Provide a brief explanation for your rating, referring to specific aspects of the response and the conversation.
\\
\\
Example:

Conversation History: 

\{\{Document\}\}

Corresponding Fact:

\{\{Fact\}\}

Response:

\{\{Response\}\}
\\
\\
Evaluation Form (scores ONLY):

- Engagingness:
}
\end{quote}

\paragraph{Evaluate Hallucinations}

\begin{quote}
{\itshape
Human Evaluation of Text Summarization Systems:
\\
\\
Factual Consistency: Does the summary untruthful or misleading facts that are not supported by the source text?
\\
\\
Source Text: 

\{\{Document\}\}

Summary: 

\{\{Summary\}\}
\\
\\
Does the summary contain factual inconsistency? 

Answer: }
\end{quote}

\end{document}